\documentclass{article}
\usepackage{spconf,amsmath,graphicx}
\usepackage{amsmath,amssymb} 
\usepackage{color}
\usepackage{tabularx}

\usepackage{times}
\usepackage{epsfig}
\usepackage{graphicx, subfigure}
\usepackage{amsmath}
\usepackage{amssymb}
\usepackage{algorithm}
\usepackage[noend]{algpseudocode}
\usepackage{multirow}
\usepackage{dsfont}

\usepackage{array, boldline, makecell, booktabs}

\usepackage[svgnames, table]{xcolor}

\title{Localization Uncertainty-Based Attention for object Detection}
\name{Sanghun Park\qquad Kunhee Kim\qquad Eunseop Lee\qquad Daijin Kim}
\address{Dept. of Computer Science and Engineering, Pohang University of Science and Technology, Korea} 
\begin{document}
%
%
\maketitle
\begin{abstract}
Object detection has been applied in a wide variety of real world scenarios, so detection algorithms must provide confidence in the results to ensure that appropriate decisions can be made based on their results.
Accordingly, several studies have investigated the probabilistic confidence of bounding box regression. 
However, such approaches have been restricted to anchor-based detectors, which use box confidence values as additional screening scores during non-maximum suppression (NMS) procedures. 
In this paper, we propose a more efficient uncertainty-aware dense detector (UADET) that predicts four-directional localization uncertainties via Gaussian modeling.  
Furthermore, a simple uncertainty attention module (UAM) that exploits box confidence maps is proposed to improve performance through feature refinement.
Experiments using the MS COCO benchmark show that our UADET consistently surpasses baseline FCOS, and that our best model, ResNext-64x4d-101-DCN, obtains a single model, single-scale AP of 48.3$\%$ on COCO test-dev, 
thus achieving the state-of-the-art among various object detectors.

\end{abstract}
%
\begin{figure}[h!]
  \begin{minipage}[h]{1.0\linewidth}
    \centering
    \centerline{\includegraphics[width=7.3cm]{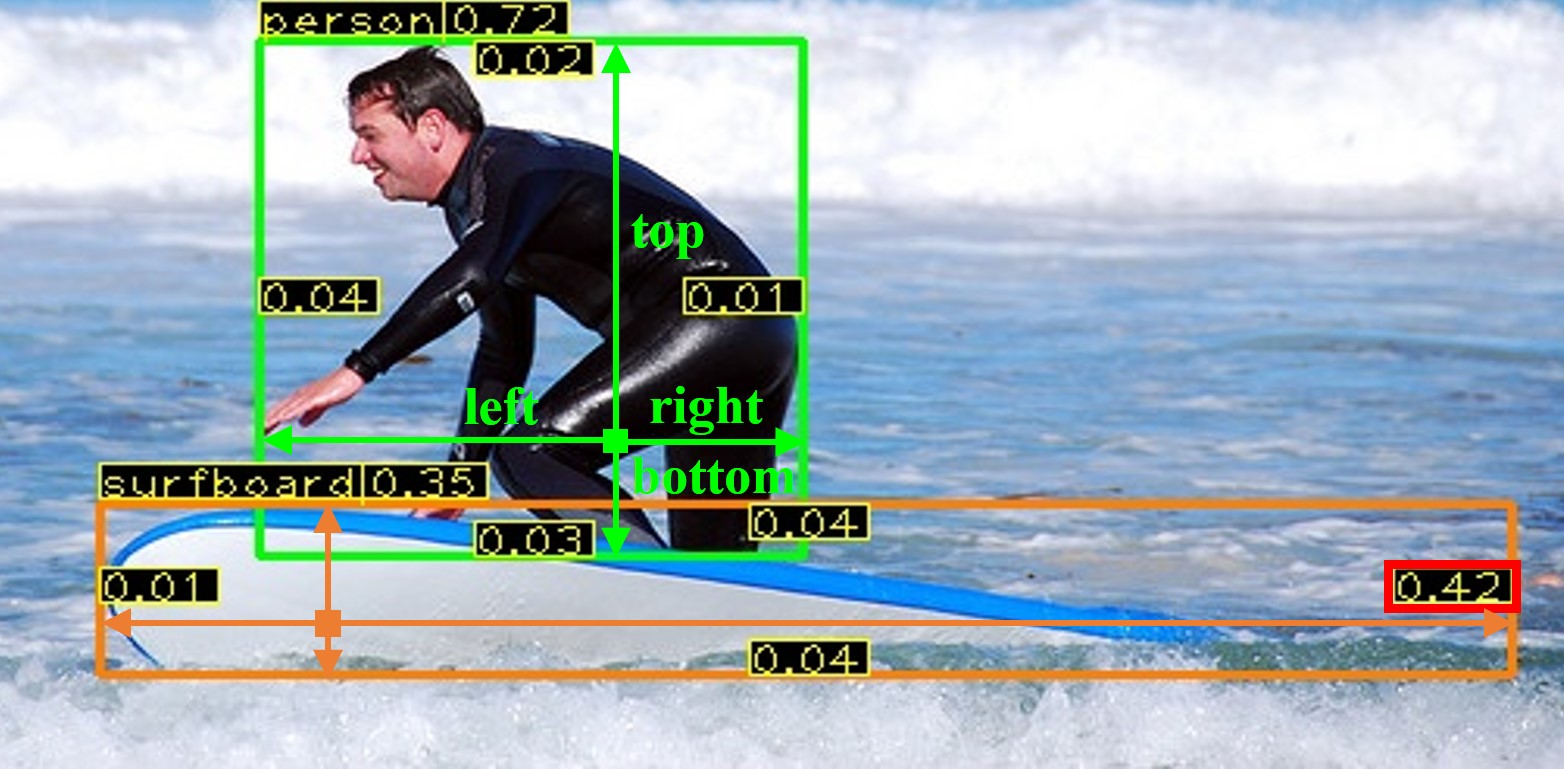}}
    \centerline{(a)}\medskip
  \end{minipage}
  \begin{minipage}[h]{1.0\linewidth}
    \centering
    \centerline{\includegraphics[width=7.3cm]{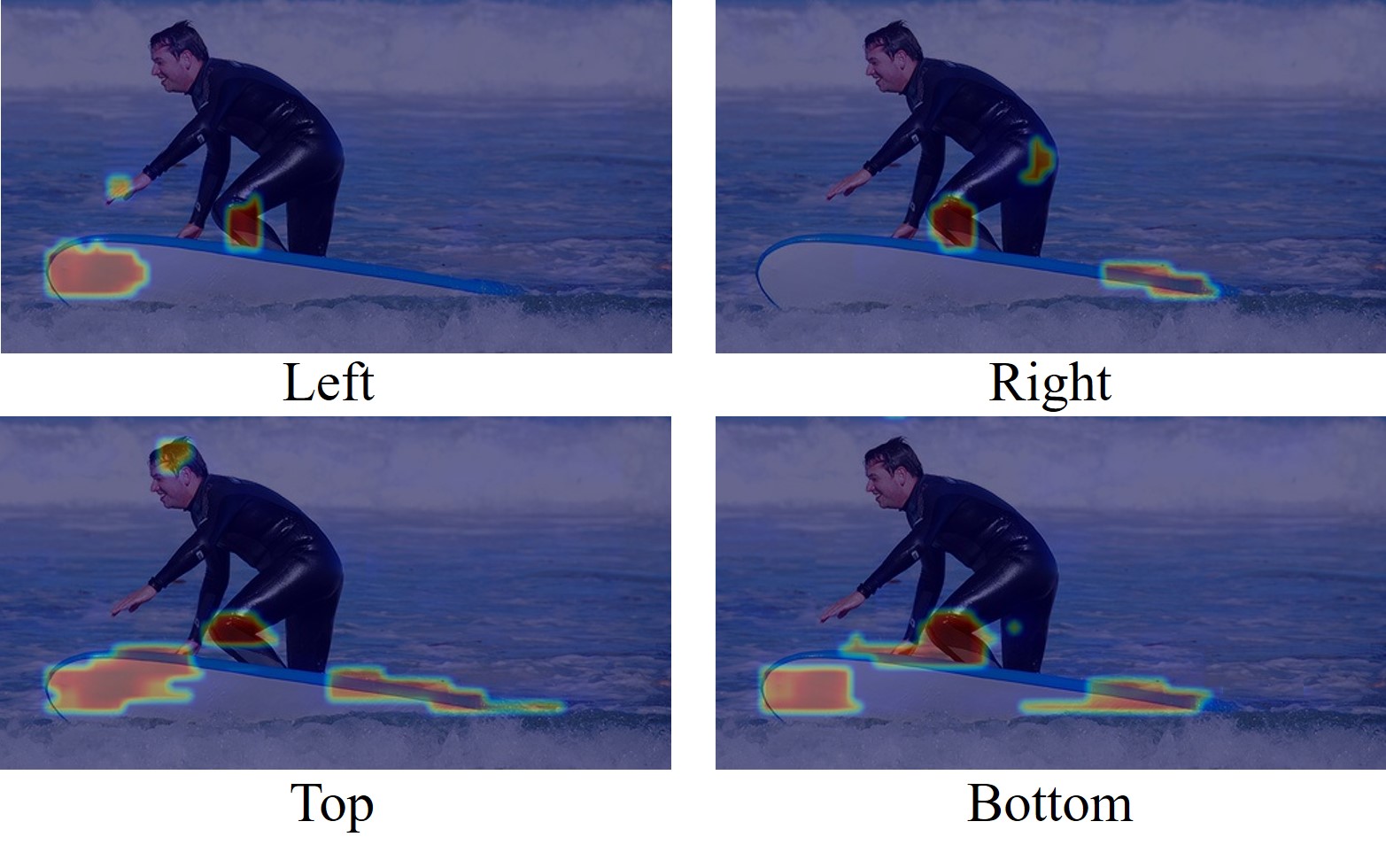}}
  \vspace{-1.3em}
    \centerline{(b)}\medskip
  \end{minipage}
  \vspace{-1.5em}
  \caption{
  The detection result of the baseline FCOS + localization uncertainty prediction.
  (a) Four-directional localization uncertainties of each detection instance.
  (b) Box confidence maps indicating the most convincing area for each boundary.}
  \label{fig:visualization}
\vspace{-1.0em}
\end{figure}
\vspace{-0.2em}
\section{Introduction}
\vspace{-0.2em}
\label{sec:intro}
Object detection has achieved noteworthy successes with the development of convolutional neural networks (CNNs).
In recent years, FCOS \cite{tian2019fcos}, an anchor-free approach, has been suggested as a strong alternative to anchor-based mainstream methods \cite{ren2016faster,lin2017focal}.
The per-pixel fashion framework achieves improved performance by modeling each object as dense key points, but it still has insufficient localization information in terms of probabilistic confidence.
Inspired by recent works on the bounding box (bbox) regression with uncertainty \cite{he2019bounding, choi2019gaussian},
we construct a localization uncertainty prediction based on FCOS \cite{tian2019fcos}
to identify the locations of the most convincing key points. 
The dense key points-based model can learn the localization uncertainty from a Gaussian model parameterized with bbox regression values and corresponding variances.

The results are shown in Fig. \ref{fig:visualization} (a), where it can be seen that the right side of the surfboard has relatively high
uncertainty because it is occluded by the water; in other words, there is a large gap between the prediction and the ground truth for the submerged part of the surfboard (marked in red). 
Fig. \ref{fig:visualization} (b) shows box confidence maps highlighting areas with key points that have low localization uncertainties for each boundary of the bboxes (highlighted in red).
As shown in these maps, it is difficult to discover the common convincing key point for the four-directional boundaries of the surfboard, while this is not the case for the person. 
Therefore, the key point with the inaccurate prediction must be chosen as the final key point for the surfboard.



This inaccurate prediction needs to be compensated by the convincing features for the boundaries of the bbox, which would be missed in the detection process of a conventional dense key points-based detector.
To this end, the localization uncertainty-based attention is designed to encode features from both the convincing regions for the boundaries of the bbox and the central region of the object.
It uses the box confidence maps to enhance the original features by exploiting certain boundary features.
Such features also allow us to effectively refine the coarse predictions.

To verify the effectiveness of the proposed method, we build an uncertainty-aware dense detector (UADET) that learns the localization uncertainty and leverages box confidence maps as spatial attentions for feature refinement; 
we then evaluate the proposed UADET on the MS COCO \cite{lin2014microsoft} benchmark. 
The experimental results show that our approach can achieve a single model, single-scale AP of 48.3$\%$, thereby representing a large improvement over the baseline FCOS \cite{tian2019fcos}.
Further, the results of extensive experiments demonstrate that the UADET significantly outperforms various state-of-the-art object detectors.
\vspace{-0.5em}
\section{Proposed Method}
\label{sec:method}
\subsection{Localization Uncertainty}
In Gaussian YOLOv3 \cite{choi2019gaussian}, due to its anchor-based design, localization uncertainty is modeled using the center point of each box, the box size, and the corresponding variances as Gaussian parameters.
In this paper, localization uncertainty is modeled using each single Gaussian model of the bbox regression values ($l$, $r$, $t$, $b$) as well as the corresponding variances. 
A single variate Gaussian distribution is as follows:
\begin{align} 
  \begin{split}
  P(x) = \cfrac{1}{\sqrt{2\pi\sigma^2}}e^{-\frac{(x - \mu)^2}{2\sigma^2}}
  \label{eq0}
  \end{split}					
\end{align}
where $\mu$ denotes the predicted bbox regression, $x$ denotes the bbox regression target, and $\sigma$ (standard deviation) denotes the localization uncertainty, the value of which is (0, 1) with a sigmoid function. 
For training, we design a negative log likelihood (NLL) loss with Gaussian parameters as follows:
\begin{align} 
  \begin{split}
  L_{Gaussian} = -\cfrac{\lambda}{N_{pos}}{\sum_{x,y}}{\:}{\sum_{l,r,t,b}}{\mathds{1}_{\{c^{G}_{x,y}{{>}0}\}}}{\log}(P(\mathds{X}))
  \label{eq1}
  \end{split}					
  \end{align}
where $\mathds{X}$ denotes $(l^G, r^G, t^G, b^G)$ of four-directional bbox regression targets, and the corresponding Gaussian parameters $\mu$ and $\sigma$ in Eq. (\ref{eq0}) are also $(\mu_l, \mu_r, \mu_t, \mu_b)$ and $(\sigma_l, \sigma_r, \sigma_t, \sigma_b)$, respectively.
$\mathds{1}_{\{c^{G}_{x,y}{{>}0}\}}$ is the indicator function, which is 1 if $c^G_{x,y}>0$ and 0 otherwise, and $c^G_{x,y}$ denotes the classification label at the $(x,y)$ pixel location of the feature. 
The summation is calculated over four-directional bbox regressions and positive samples. 
The cost average is calculated by dividing by the number of positive samples, $N_{pos}$. 
$\lambda$ ($\lambda\,$= 0.2 in this paper) is the balance weight for $L_{Gaussian}$.

In contrast to Gaussian YOLOv3 \cite{choi2019gaussian}, we observe degraded performance when bbox regression is learned solely from $L_{Gaussian}$.  
Thus, we set the $\lambda$ value of $L_{Gaussian}$ such that the impact on bbox regression will be less than the powerful GIoU \cite{rezatofighi2019generalized} loss, but enough to learn the standard deviation.
According to $L_{Gaussian}$, the localization uncertainty is predicted to involve larger $\sigma$ values when there are larger gaps between the predicted regressions and corresponding targets, and vice versa. 
Therefore, we can utilize ($1.0 - uncertainty$) as the four-directional box confidence at each pixel location. 
%
\subsection{Uncertainty Attention Module}
The dense key points-based detector typically focuses on the area at the center of the object, as this usually ensures powerful feature representation for predictions.
Accordingly, the pixel location with the maximum classification score within the central area is selected as the final key point location for initial predictions.
However, we can check that the convincing regions for the boundaries of the bbox maintain strong representation for bbox regression from the obtained box confidence maps (see Fig. \ref{fig:visualization} (b)).
Occasionally, such features can also be better representations for classification than the center feature of the object in cases of occlusion by the background or unusually shaped objects.
This means that we can compensate for initial predictions focusing on the center region of the object by exploiting the convincing features for the boundaries of the bbox, indicated by localization uncertainty. 

\begin{figure*}[t]
	\begin{center}
    \includegraphics[width=0.96\linewidth]{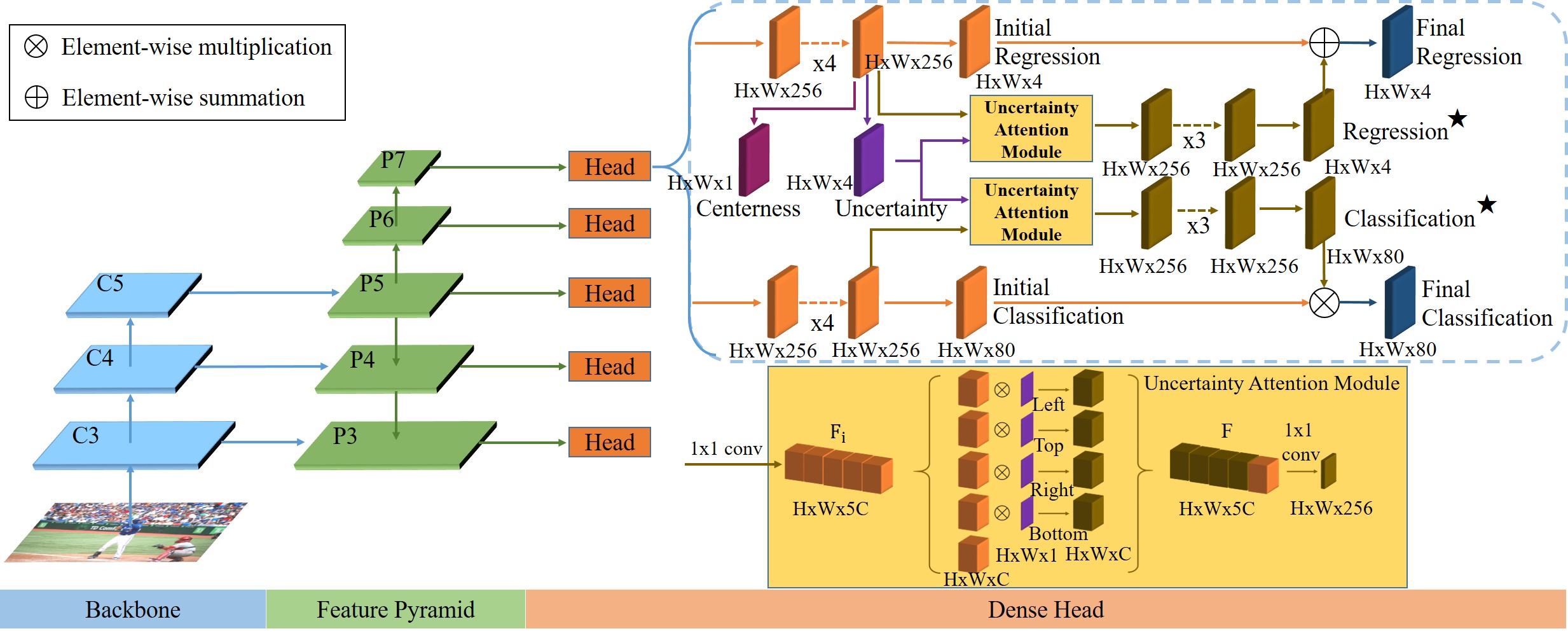}
  \end{center}
  \vspace{-2em}
  \caption{Network architecture of the uncertainty-aware dense detector (UADET).}
  \vspace{-0.5em}
	\label{fig:network}
\end{figure*}

We thus propose a novel feature refinement module, the uncertainty attention module (UAM) that leverages box confidence maps as spatial attentions. 
As shown in Fig. \ref{fig:network}, the UAM takes the last feature of the initial prediction as input,
then generates a feature $F_i$ with (4+1)C channels through the 
$1\times1$ convolution layer. The 4C channels of $F_i$ correspond to the box confidence map of each boundary, while the other 1C channels of $F_i$ correspond to the original feature representing the central area of the object.
Each box confidence map is multiplied spatially to the corresponding $F_i$, then all the features are concatenated. The concatenated feature $F$ can be formulated using the following equation:
\begin{align} 
\begin{split}
F = 
\left\{ 
  \renewcommand\arraystretch{1.2}
  \renewcommand{\arraycolsep}{1.0mm}
  \begin{array}{llrl} 
     F_{ic} \otimes (1.0 - U(L))&, &\quad 0\leq &c<C\\
     F_{ic} \otimes (1.0 - U(T))&, &\quad C\leq &c<2C\\
     F_{ic} \otimes (1.0 - U(R))&, &\quad 2C\leq &c<3C\\
     F_{ic} \otimes (1.0 - U(B))&, &\quad 3C\leq &c<4C\\
     F_{ic}                     &, &\quad 4C\leq &c<5C\\
  \end{array} 
\right.
\label{eq2}
\end{split}
\end{align}
%
where $c$ denotes the feature channel and $U(L)$, $U(T)$, $U(R)$ and $U(B)$ respectively denote the localization uncertainties of the left, top, right and bottom.
Finally, the UAM produces an output feature with the same shape as the input feature through a $1\times1$ convolution layer.
In this paper, we apply C = 256 for the classification refinement branch and C = 64 for the bbox regression refinement branch.
\vspace{-0.5em}
\subsection{Network Architecture}
Fig. \ref{fig:network} illustrates the overall network architecture of the uncertainty-aware dense detector (UADET).
The structures of the backbone and the feature pyramid network (FPN) are the same as those of the FCOS \cite{tian2019fcos}, but the head structure is different. 
First, we attach the localization uncertainty prediction to the initial bbox regression branch. 
Then, we attach additional sub-branches for classification and bbox regression refinement using the UAM, which leverages localization uncertainty.
Each sub-branch refines the feature through the UAM, and finally applies $3\times3$ convolution layers to produce the prediction to be refined. 
The UADET predicts final classification and bbox regression by combining the existing and refined results.
%
\vspace{-0.8em}
\subsection{Loss Function}
\vspace{-0.2em}
We model the sub-branches as a generated anchor refinement problem. 
We serve the initial bbox prediction as an anchor generated from the pixel location of the feature.
We then obtain the classification label by measuring the intersection over unit (IoU) between the generated anchor and the ground truth boxes. 
The classification label of the ground truth box, which has a maximum IoU with the generated anchor, is the label of the anchor. If the maximum IoU is under 0.6, we treat that anchor as the background.
We adopt focal loss \cite{lin2017focal} for classification refinement branch. 
\vspace{-0.2em}
\begin{align} 
  \begin{split}
    L_{cls^\star} = \cfrac{1}{N_{pos^{\star }}}{\sum_{x,y}}{L_{focal}}{(p^{\star}_{x,y},\,c^{\ast}_{x,y})}
    \label{eq3}
  \end{split}					
\end{align}
$N_{pos^{\star}}$ denotes the number of positive samples from the above classification targeting strategy while $p^{\star}_{x,y}$, $c^{\ast}_{x,y}$ respectively denote the classification score and target for refinement.

For positive samples, the generated anchor is compensated by the offset to the assigned ground truth box.
The sub-branch for bbox regression refinement learns the offset through L1 loss. 
\vspace{-0.5em}
\begin{align} 
  \begin{split}
    L_{reg^\star} = \cfrac{1}{N_{pos^{\star }}}{\sum_{x,y}}{\mathds{1}_{\{c^{\ast}_{x,y}{{>}0}\}}}{\parallel}{\delta_{x,y},\,\delta^{\ast}_{x,y}}{\parallel}
    \label{eq4}
  \end{split}					
\end{align}
$\delta_{x,y}$, $\delta^{\ast}_{x,y}$ respectively denote the four-directional bbox regression offset and the corresponding target.
$\mathds{1}_{\{c^{\ast}_{x,y}{{>}0}\}}$ is the indicator function, which is 1 if $c^{\ast}_{x,y}>0$ and 0 otherwise.
Finally, we define the total training loss as follows:
\begin{align} 
  \begin{split}
  L_{Total} = L_{cls} &+ L_{reg} + L_{centerness}\; + \\
      L_{cls^\star} &+ L_{reg^\star} + L_{Gaussian}
    \label{eq5}
  \end{split}					
\end{align}
Following FCOS \cite{tian2019fcos}, we adopt focal loss \cite{lin2017focal} for initial classification, GIoU loss \cite{rezatofighi2019generalized} for initial regression, and binary cross-entropy loss for centerness.
\vspace{-0.5em}
\subsection{Inference}
The inference of the UADET is straightforward: First, we forward an input image through the network and obtain the
initial classification score $p_{x,y}$, initial bbox regression $t_{x,y} = (l, r, t, b)$, centerness $c_{x,y}$, localization uncertainty $u_{x,y}$, 
classification score for refinement $p^{\star}_{x,y}$, and bbox regression offset $\delta_{x,y} = (\delta_l, \delta_r, \delta_t, \delta_b)$ 
for each $(x,y)$ pixel location of the feature.
Next, we adopt the final classification score as the square root of $p_{x,y}$ ${\times}$ $p^{\star}_{x,y}$ and the final bbox regression as the summation of $t_{x,y}$ and $\delta_{x,y}$. 
Then, we process centerness-weighted NMS with a threshold of 0.6 to eliminate redundant detections.
%
\vspace{-0.5em}
\section{Experiments}
\label{sec:pagestyle}
We evaluate our UADET on the large-scale detection benchmark MS COCO \cite{lin2014microsoft}. Following common practice \cite{tian2019fcos, lin2017focal, zhang2020atss},
we use the train2017 split for training and report the ablation results on the val2017 split. To compare ours with state-of-the-art
detectors, we report our main results on test-dev split by uploading our detection results to the evaluation server.
\begin{table*}[t]
  \renewcommand{\tabcolsep}{4.1mm}
  \renewcommand{\arraystretch}{0.90}
  \caption{Detection results($\%$) on MS COCO test-dev split.}
  {\small
	\begin{center}
		\begin{tabular}{l | c | c c c | c c c}
			\Xhline{1pt}
			Method & Backbone  & AP & $AP_{50}$ & $AP_{75}$ & $AP_S$ & $AP_M$ & $AP_L$  \\
			\hline
      \hline
      Faster R-CNN w/ FPN \cite{lin2017FPN} &		ResNet-101 &	36.2 &	59.1 &	39.0 &	18.2 &	39.0 &	48.2 \\
      Mask R-CNN \cite{he2017mask}          &		ResNet-101 &	38.2 &	60.3 &	41.7 &	20.1 &	41.1 &	50.2 \\
      Cascade R-CNN \cite{cai2018cascade}   &		ResNet-101 &	42.8 &	62.1 &	46.3 &	23.7 &	45.5 &	55.2 \\
      \hline
      DSSD513 \cite{fu2017dssd}                &	ResNet-101 &	33.2 &	53.3 &	35.2 &	13.0 &	35.4 &	51.1 \\
      RetinaNet \cite{lin2017focal}            & ResNet-101 &	39.1 &	59.1 &	42.3 &	21.8 &	42.7 &	50.2 \\
      RefineDet512+ \cite{zhang2018single}     & ResNet-101 &	41.8 &	62.9 &	45.7 &	25.6 &	45.1 &	54.1 \\
      FreeAnchor \cite{zhang2019freeanchor}    & ResNet-101 &	43.1 &	62.2 &	46.4 &	24.5 &	46.1 &	54.8 \\
      \hline
      CornerNet \cite{law2018cornernet}      & Hourglass-104  &	40.5 &	56.5 &	43.1 &	19.4 &	42.7 &	53.9 \\
      CenterNet \cite{duan2019centernet}     & Hourglass-104  &	44.9 &	62.4 &	48.1 &	25.6 &	47.4 &	57.4 \\
      RepPoints \cite{yang2019reppoints}     & ResNet-101-DCN &	45.0 &	66.1 &	49.0 &	26.6 &	48.6 &	57.5 \\
      \hline
      FSAF \cite{zhu2019fsaf}          & ResNet-101 &	40.9 &	61.5 &	44.0 &	24.0 &	44.2 &	51.3 \\
      ATSS \cite{zhang2020atss}        & ResNet-101 &	43.6 &	62.1 &	47.4 &	26.1 &	47.0 &	53.6 \\
      FCOS \cite{tian2019fcos}         & ResNeXt-64x4d-101 &	44.7 &	64.1 &	48.4 &	27.6 &	47.5 &	55.6 \\
      \hline
      UADET $\textit{(ours)}$  & ResNet-101            &	44.0 &	62.6 &	47.7 &	26.1 &	47.1 &	54.5 \\
      UADET $\textit{(ours)}$  & ResNeXt-64x4d-101     &	46.0  &	65.0  &	50.0  &	28.6  &	48.9  &	56.8  \\
      UADET $\textit{(ours)}$  & ResNet-101-DCN        &	46.4  &	65.1  &	50.5  &	27.7  &	49.4  &	58.6  \\
      UADET $\textit{(ours)}$  & ResNeXt-64x4d-101-DCN &	\textbf{48.3}  &	\textbf{67.2}  &	\textbf{52.5}  &	\textbf{30.1}  &	\textbf{51.2}  &	\textbf{61.0}  \\
			\Xhline{1pt}
    \end{tabular}
    \vspace{-1.0em}
  \end{center}
  }
	\label{maintable}
\end{table*}
\begin{table}
  \renewcommand{\tabcolsep}{1.0mm}
  \renewcommand{\arraystretch}{1.0}
  \vspace{-2em}
  \caption{Individual component contributions.}
  \begin{center}
		\begin{tabular}{l | c c c | c c c}
			\Xhline{1pt}
			Method & AP & $AP_{50}$ & $AP_{75}$ & $AP_{S}$ & $AP_{M}$ & $AP_{L}$ \\
			\hline
			\hline
      FCOS\cite{tian2019fcos}   & 38.6  &  57.4 & 41.4 & 22.3 & 42.5 & 49.8 \\
      \hline
			+ Uncertainty   & 38.7  &  56.9 & 41.6 & 22.6 & 42.2 & 50.4 \\
      + Cls. refinement    & 39.3  &  57.5 & 43.0 & 23.2 & 43.1 & \textbf{52.1} \\
      + Reg. refinement    & \textbf{39.9}  &  \textbf{58.1} & \textbf{43.2} & \textbf{23.3} & \textbf{43.7} & 51.8 \\
			\Xhline{1pt}
    \end{tabular}
    \vspace{-0.2em}
  \end{center}
	\label{abltable}
\end{table}
\begin{table}
  \renewcommand{\tabcolsep}{1.2mm}
  \renewcommand{\arraystretch}{1.0}
  \vspace{-1.9em}
  \caption{Contribution of the UAM. The UAM is replaced with a 1$\times$1 convolution layer in the first row.}
  \vspace{-0.8em}
  \begin{center}
		\begin{tabular}{l | c | c c c | c c c}
			\Xhline{1pt}
			Method & UAM & AP & $AP_{50}$ & $AP_{75}$ & $AP_{S}$ & $AP_{M}$ & $AP_{L}$ \\
			\hline
			\hline
      UADET         & & 39.1 & 57.7 & 42.3 & 22.9 & 42.6 & 51.1 \\
      \hline
      UADET         &\checkmark & \textbf{39.9} & \textbf{58.1} & \textbf{43.2} & \textbf{23.3} & \textbf{43.7} & \textbf{51.8} \\    
			\Xhline{1pt}
    \end{tabular}
    \vspace{-2.2em}
  \end{center}
	\label{abltable3}
\end{table}
\vspace{-0.5em}
\subsection{Implementation and Training Details}
We implement the UADET based on MMDetection \cite{chen2019mmdetection}. Unless specified, the default hyper-parameters of FCOS \cite{tian2019fcos} implementation are maintained.
For the ablation results, we adopt ResNet-50 pretrained on ImageNet as the backbone. The network is trained with an initial learning rate of 0.01 and a mini-batch size of 16 for 90K iterations.
The input images are resized to a scale of 1333 $\times$ 800 without changing the aspect ratio, then augmented by random flipping. 

To allow for a fair comparison with the state-of-the-art detectors, following prior work on FCOS \cite{tian2019fcos}, we adopt larger backbone networks and a multi-scale training strategy. 
The shorter side of the input images are randomly resized in the range of [640:800], and the number of training iterations is doubled to 180K.
At testing time, we adopt a single-scale testing strategy for all results.
%
%
\vspace{-0.7em}
\subsection{Ablation Study}
We now assess the effectiveness of each component in our method. As presented in Table \ref{abltable}, 
the performance at baseline is slightly improved with the effectiveness of $L_{Gaussian}$ when we attach the localization uncertainty prediction.  
Due to the refined features of the UAM and the IoU-based targeting strategy on the classification refinement branch, the performance is boosted up to 39.3$\%$ in AP.
By adding the regression refinement branch to compensate for initial bbox regressions, we obtain a further boosted performance of 39.9$\%$ in AP.

We also verify the effectiveness of the UAM, as presented in Table \ref{abltable3}. 
For one, the UAM outperforms a conventional convolutional layer.
This indicates that the UAM extracts more effective feature representations by considering the convincing features for the boundaries of the bbox.
\subsection{Comparison with State-of-the-art Detectors}
We compare our UADET with recent state-of-the-art detectors on the MS COCO test-dev. 
As listed in Table \ref{maintable}, the UADET achieves 44.0$\%$ in AP, thus surpassing other methods with the same ResNet-101 backbone.
With the ResNeXt-64x4d-101 backbone, the performance is further boosted to 46.0$\%$ in AP, which is a substantial improvement over the baseline performance.
By applying deformable convolution \cite{zhu2019deformable} (DCN) 
at stages 3 and 4 of the backbone and the last convolution of 4$\times$ repeated convolutions shown in Fig. \ref{fig:network},
the performance is improved to 46.4$\%$ in AP with the ResNet-101 backbone. 
Finally, our best model, which includes the ResNeXt-64x4d-101 backbone and DCN, achieves 48.3$\%$ in AP, 
thereby offering state-of-the-art performance as a single model with a single-scale testing strategy.
\vspace{-0.6em}
\section{CONCLUSIONS}
\label{sec:typestyle}
\vspace{-0.5em}
In this work, we propose a novel feature refinement method using localization uncertainty-based attention.
We then build an uncertainty-aware dense detector (UADET) that utilizes this novel method to improve performance.
The experimental results show that UADET achieves remarkable performance compared to various detectors, including recent state-of-the-art detectors.
%
\vspace{-1.0em}
\section{ACKNOWLEDGEMENT}
\label{sec:typestyle}
\vspace{-0.5em}
This work was supported by Institute of Information $\&$ communications Technology Planning $\&$ Evaluation (IITP) grant funded by the Korea government (MSIT). (No.2018-0-01290, Development of an Open Dataset and Cognitive Processing Technology for the Recognition of Features Derived From Unstructured Human Motions Used in Self-driving Cars) and (No.B0101-15-0266, Development of High Performance Visual BigData Discovery Platform for Large-Scale Realtime Data Analysis)
\bibliographystyle{IEEEbib}
\bibliography{egbib}
\end{document}